\documentclass[letterpaper, left=48pt, right=48pt, bottom=47pt, top=60pt, conference]{ieeeconf}  

\IEEEoverridecommandlockouts                              

\usepackage{times} 

\usepackage{amsmath} 
\usepackage{amssymb}  
\usepackage{amsfonts}
\usepackage{amsthm}
\usepackage{bm}
\usepackage[final]{changes}
\usepackage{graphicx}
\usepackage{epstopdf}
\usepackage{subfigure}
\usepackage{booktabs}
\usepackage[flushleft]{threeparttable}
\usepackage{multirow}
\usepackage{makecell}
\usepackage[numbers,sort&compress]{natbib}
\usepackage[linesnumbered,ruled,vlined]{algorithm2e}
\usepackage{amssymb}
\usepackage{pifont}
\usepackage{caption}
\usepackage{balance}
\usepackage[colorlinks,
            linkcolor=blue,
            anchorcolor=blue,
            citecolor=blue]{hyperref}

\theoremstyle{definition}

\title{\LARGE \bf
  GPU-Accelerated Optimization-Based Collision Avoidance
}

\author{
      Zeming~Wu$^{1}$,
      Zhuping~Wang$^{1}$,
      and Hao~Zhang$^{1}$
      \thanks{$^{1}$Z.~Wu, Z.~Wang and H.~Zhang are with the Department of Control Science and Engineering, Tongji University, Shanghai, China, e-mail: \{zemingwu, elewzp, zhang\_hao\}@tongji.edu.cn.}
}

\begin{document}

\bstctlcite{IEEEexample:BSTcontrol}

\maketitle
\thispagestyle{empty}
\pagestyle{empty}

\begin{abstract}
  This paper proposes a GPU-accelerated optimization framework for collision avoidance problems where the controlled objects and the obstacles can be modeled as the finite union of convex polyhedra.
  A novel collision avoidance constraint is proposed based on scale-based collision detection and the strong duality of convex optimization.
  Under this constraint, the high-dimensional non-convex optimization problems of collision avoidance can be decomposed into several low-dimensional quadratic programmings (QPs) following the paradigm of alternating direction method of multipliers (ADMM).
  Furthermore, these low-dimensional QPs can be solved parallel with GPUs, significantly reducing computational time.
  High-fidelity simulations are conducted to validate the proposed method's effectiveness and practicality.
\end{abstract}

\section{Introduction}
With the rapid advancement of autonomy, deploying autonomous systems in complex environments with obstacles has become increasingly prevalent.
Examples include self-driving cars navigating on urban roads \cite{wang2018collision,ding2021epsilon} and autonomous quadrotors maneuvering through densely forested areas \cite{loquercio2021learning,zhou2022swarm}.
In such scenarios, achieving high-precision collision avoidance is paramount to ensure safe navigation.

Various approaches have been developed to achieve collision avoidance, including search-based methods \cite{liu2018search}, sampling-based methods \cite{wang2020sampling}, and optimization-based methods \cite{zhang2020optimization,wang2022geometrically,han2023rda,wang2023linear}.
Search-based methods discretize the configuration space into grids with a defined resolution and systematically search these grids to find a feasible solution.
Sampling-based methods employ various sampling schemes to probe the configuration space for a feasible solution.
While search-based and sampling-based methods have found extensive applications in the real world, they encounter computational challenges when precise consideration of both robot and obstacle geometry is necessary.
This is notably evident in situations such as the intricate whole-body planning of quadrotors to navigate through narrow gaps \cite{wang2022geometrically} or overtaking in dense traffic flows \cite{wang2023linear}.

Optimization-based methods, on the other hand, achieve safe navigation by minimizing a prescribed performance index and adhering to dynamics, kinematics, and collision avoidance constraints.
Recent research \cite{zhang2020optimization,wang2022geometrically,wang2023linear,han2023rda} has demonstrated the advantages of optimization-based approaches, particularly in the context of high-precision collision avoidance.
Furthermore, optimization-based methods, such as model predictive control (MPC) \cite{zhang2020optimization,han2023rda} and trajectory optimization \cite{wang2022geometrically,wang2023linear} are inherently capable of finding better trajectories under certain performance index.
Despite these promising features, there are three challenges hindering the development of optimization-based methods in real-world applications.
Firstly, the precise formulation of collision avoidance constraint is hard to handle in general optimization frameworks.
Secondly, the dimension of the optimization problem increases considerably with the number of obstacles.
Thirdly, the optimization problems are generally non-convex due to the system dynamic constraints and collision avoidance constraints.
These features make the optimization-based methods hard to achieve real-time navigation on embedded devices.

To address these challenges, this paper proposes a novel optimization-based framework for the collision avoidance.
The geometry of controlled objectives and obstacles are modeled as finite unions of polyhedra, which can satisfy the precision requirements in most scenarios.
Furthermore, the proposed framework can fully exploit the optimization problem's inherent structure, leveraging GPUs' power to expedite the solving process.
The main contributions are summarized as follows.
\begin{itemize}
  \item A novel scale-based collision avoidance constraint for polyhedra is proposed.
        Compared with the widely used signed distance-based constraint \cite{zhang2020optimization}, the proposed constraint is linear with respect to dual variables.
  \item Leveraging the linear characteristic of the scale-based collision avoidance constraint, we break down the high-dimensional non-convex optimization problem of collision avoidance into multiple low-dimensional QPs following the paradigm of ADMM.
        These QPs can be solved parallel with GPUs, resulting in a significant reduction in computation time.
  \item High-fidelity simulations are conducted to evaluate the effectiveness and practicality of our framework on embedded platforms.
\end{itemize}

\vspace{-0.3em}
\section{Related Works}
In this section, a brief literature review of the optimization-based methods for collision avoidance is presented.
\vspace{-0.4em}
\subsection{Collision Avoidance Constraints}
The appropriate formulation of collision avoidance constraints is crucial for the computational overhead of optimization-based collision avoidance.
In \cite{mellinger2012mixed}, collision avoidance constraints are established through the identification of a separating plane that distinguishes the quadrotor's position from obstacles on opposing sides.
This approach leads to an optimization problem that involves integer variables, which takes several seconds to solve.
In \cite{liu2017planning} the free space is decomposed into a series of convex overlapping polyhedra, namely safe flight corridors (SFCs), and then these polyhedra constraints are impose to optimization problem for generating collision-free trajectories.
Nevertheless, SFCs can lead to conservative solutions due to the oversimplification of free space.
For the applications requiring precise consideration of both robot and obstacle geometry, signed distance constraint \cite{schulman2014motion,zhang2020optimization,wang2022geometrically} are commonly used.
However, the signed distance constraint is implicit and non-differentiable, which makes it hard to handle in general optimization frameworks.
Recently, a scale-based collision avoidance constraint has been proposed for convex set in \cite{tracy2023differentiable}.
Although this constraint is piecewise differentiable, it still retains an implicit and non-smooth nature.
Moreover, calculating the gradient involves conic programming, which is computationally expensive on embedded platforms.
To this end, an explicit, smooth and easily manageable constraint for collision avoidance is proposed in this paper.


\subsection{Collision Avoidance with Optimization}
Various approaches have been proposed to achieve collision avoidance with optimization.
In works such as \cite{kalakrishnan2011stomp,zucker2013chomp,wang2022geometrically,han2021fast,wang2023linear}, the problem of collision-free trajectories generation is formulated as an unconstrained optimization problem.
Within these frameworks, collision avoidance constraints are translated into penalty functions.
However, these frameworks are bothered by unavoidable constraint violations, particularly in narrow, dense environments.
Constrained optimization is introduced in \cite{mellinger2012mixed,schulman2014motion,zhang2020optimization,han2023rda}.
Unfortunately, it takes seconds to minutes to solve mixed-integer quadratic programming \cite{mellinger2012mixed} or constrained nonlinear programming \cite{schulman2014motion,zhang2020optimization}, which makes them unsuitable for real-time navigation.
As an attempt to enhance real-time performance, the high-dimensional non-convex optimization problem from \cite{zhang2020optimization} is converted to multiple low-dimensional second-order conic programmings (SOCPs) in \cite{han2023rda}, whose number is proportional to the number of obstacles and prediction horizon.
Nevertheless, unlike QPs, there is no simplex or simplex-like solving method for SOCPs, making the method difficult to apply in practice.
Worse, solving so many SOCPs with CPU remains computationally expensive on embedded platforms.
In this paper, we tackle this challenges by separating the origin optimization problem into multiple low-dimensional QPs and leverage the power of GPU to expedite the solving process.

\section{Scale Based Collision Avoidance \\ Constraints of Polyhedra}

In this section, the scale-based collision detection is introduced.
Moreover, a novel scaled-based collision avoidance constraint is proposed using the strong duality of linear programming (LP).
\vspace{-0.5em}
\subsection{Scale-based Collision Detection of Polyhedra}
\label{subsec:scale-based collision detection}
The geometry of obstacles and robots are both described by polyhedra.
Specially, $A \in \mathbb{R}^{n_r \times 3}$ and $b \in
  \mathbb{R}^{n_r}$ are used to describe the robots:
\begin{equation}
  A \bm{x} \preceq  \bm{b}.
\end{equation}
$C \in \mathbb{R}^{n_o \times 3}$ and $d \in \mathbb{R}^{n_o}$ are used to describe the obstacles:
\begin{equation}
  C \bm{x} \preceq \bm{d}.
\end{equation}

\begin{figure}[t]
  \centering
  \includegraphics[width=0.5\hsize]{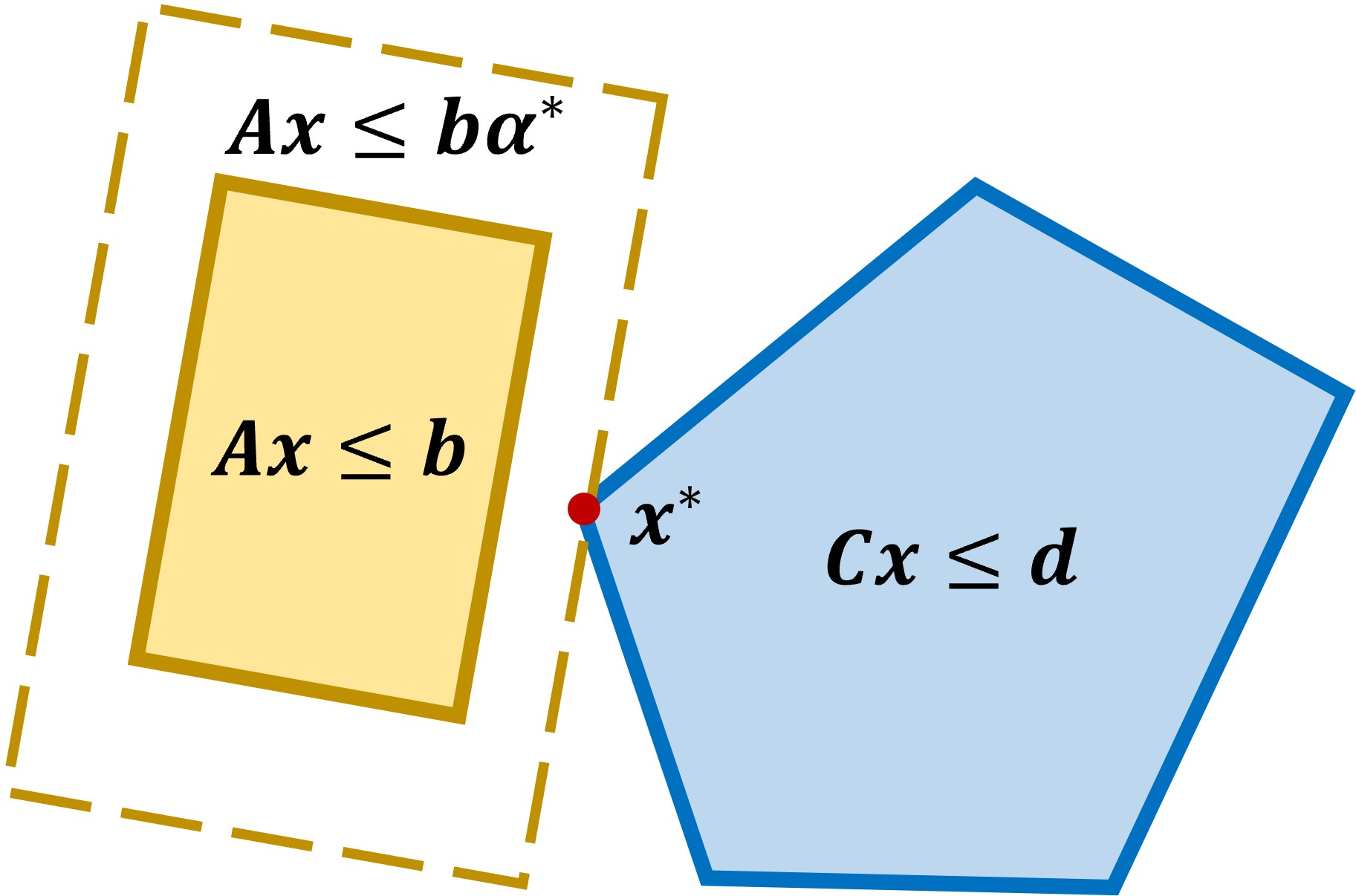}
  \caption{An illustration of scale-based collision detection.}
  \label{fig:polyhedron_collision}
  \vspace{-0.7cm}
\end{figure}

To detect whether there is collision between robot and obstacles, we need to
check whether the polyhedra of robots and obstacles have overlap.
One of the
method to detect the overlap between polyhedra is proposed in
\cite{tracy2023differentiable}.
This method enlarges/shrinks the
polyhedra of both robot and obstacle to find a smallest scale $\alpha^* \in \mathbb{R}_+$ that cause a collision.
If $\alpha^* < 1$, there is a collision between polyhedra.
On the contrary, if $\alpha^* \geq 1$, there is no collision.
In this paper, we only scale the polyhedra of robot, a geometry illustration is shown in Fig. \ref{fig:polyhedron_collision}.
The collision detection problem between robot and obstacles is formulated as the following LP:
\begin{equation}
  \label{eqn:scale_opt}
  \begin{aligned}
    \min_{\alpha,~\bm{x}} & ~~~~~~~~~~~~~~~~~\alpha                                              \\
    s.t.                  & ~~~~A \bm{x} \preceq \bm{b} ~ \alpha, \quad C \bm{x} \preceq \bm{d}.
  \end{aligned}
\end{equation}

To enforce collision-free, we must impose the constraint $\alpha^* \geq 1$ into the optimization problem.
This constraint is implicit and non-smooth, which is hard to handle in general optimization frameworks.
Consequently, it is essential to reformulate this constraint into an explicit and smooth form.

\vspace{-0.3em}
\subsection{Reformulation of Collision Constraints}
\label{subsec:reformulation}
We convert the collision avoidance constraint to an explicit one by using the strong
duality of LP.
The dual Lagrangian function of the LP (\ref{eqn:scale_opt}) is:
\begin{equation}
  \begin{aligned}
     & g(\bm{\lambda}, \bm{\mu})  = \inf_{\alpha, \bm{x}} \mathcal{L}  (\alpha, \bm{x}, \bm{\lambda}, \bm{\mu})                                                                         \\
     & = \inf_{\alpha, \bm{x}} \bigg\{\alpha + \bm{\lambda}^T(A \bm{x} - \bm{b} \alpha) + \bm{\mu}^T(C \bm{x}- \bm{d}) \bigg\}                                                          \\
     & = \inf_{\alpha} \!\! \bigg\{(1-\bm{\lambda}^T \bm{b}) \alpha\bigg\} \! + \! \inf_{\bm{x}} \!\! \bigg\{(\bm{\lambda}^T A \! + \! \bm{\mu}^T C) x\bigg\} \! - \! \bm{\mu}^T\bm{d},
  \end{aligned}
\end{equation}
where $\bm{\lambda}$ and $\bm{\mu}$ are the Lagrangian multiplier with proper dimension.
Accordingly, the dual problem of LP (\ref{eqn:scale_opt}) is:
\begin{equation}
  \label{eqn:scale_opt_dual}
  \begin{aligned}
    \max_{\bm{\lambda},~\bm{\mu}} & ~~~~~~~~~~~~~~~  -\bm{d}^T \bm{\mu}                               \\
    s.t.                          & ~~~~~~ -b^T\bm{\lambda} + 1 = 0,~~ \bm{\lambda} \succeq 0,        \\
                                  & ~~~~~~ A^T \bm{\lambda} + C^T \bm{\mu} = 0,~~ \bm{\mu} \succeq 0. \\
  \end{aligned}
\end{equation}
Due to the strong duality of LP, the objective function of (\ref{eqn:scale_opt_dual}) and (\ref{eqn:scale_opt}) have the following relationship:
\begin{equation}
  -\bm{d}^T \bm{\mu} \leq -\bm{d}^T \bm{\mu}^* = \alpha^*.
\end{equation}
Consequently, the collision-free constraint can be explicitly formulated as:
\begin{equation}
  \label{eqn:poly_colli_free_conts}
  \begin{aligned}
    -\bm{d}^T \bm{\mu} \geq 1, \qquad \bm{\lambda} \succeq 0, \qquad \bm{\mu} \succeq 0, \\
    -b^T\bm{\lambda} + 1 = 0, \qquad A^T \bm{\lambda} + C^T \bm{\mu}  = 0.               \\
  \end{aligned}
\end{equation}
It is worth pointing out that (\ref{eqn:poly_colli_free_conts}) are linear constraints with respect to $(\bm{\lambda},~\bm{\mu})$.
This feature makes it possible to decompose the optimization problem of collision avoidance into QPs.

\section{Collision Avoidance Based on Model Predictive Control}
In this section, we employ the constraint (\ref{eqn:poly_colli_free_conts}) within the framework of MPC to formulate an optimization problem.
Subsequently, in Section \ref{subsec:ADMM}, the ADMM paradigm is adopted to decompose the high-dimensional non-convex optimization problem into multiple low-dimensional QPs.
Further insights into the process of solving these QPs in parallel using GPUs are presented in Section \ref{subsec:GPU}.
Finally, time complexity of the proposed method is discussed.

\vspace{-0.5em}
\subsection{Optimization Problem Formulation}
\label{subsec:MPC}
The robot state can be evaluated by the system equation:
\begin{equation}
  \label{eqn:system_eqn}
  \bm{s}_{t+1} = \bm{s}_t + f(\bm{s}_t,\bm{u}_t),
\end{equation}
where $\bm{s}_t \in \mathbf{R}^{n_s}$ is the state of robot at $t$-th step, $\bm{u}_t \in \mathbb{R}^{n_u}$ is the control input at $t$-th step.
Note that $f(\bm{s}_t,\bm{u}_t)$ is not a linear function in general.

\begin{figure}[ht]
  \centering
  \includegraphics[width=0.5\hsize]{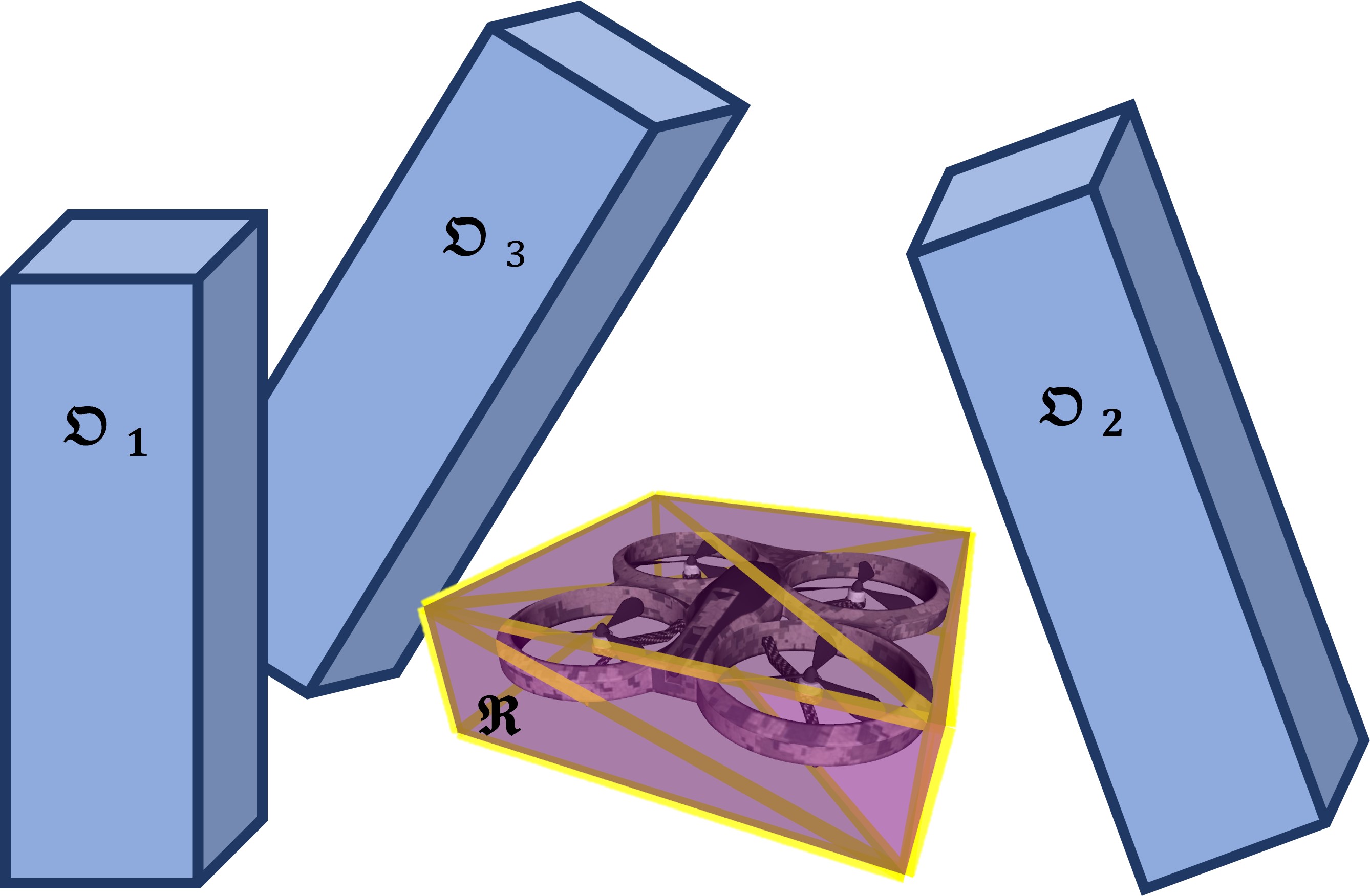}
  \caption{The geometry of quadrotor and obstacles.}
  \label{fig:geo_robot_obstacles}
  \vspace{-0.5cm}
\end{figure}

In this paper, the geometry of robot $\mathfrak{R}$ is modeled by the union of a series of compact convex set.
Each compact convex set is modeled by a polyhedron that relates to the state of robot. More precisely:
\begin{subequations}
  \begin{gather*}
    \mathfrak{R} = \mathfrak{R}_1 \cup \mathfrak{R}_2 \cup \cdots \cup \mathfrak{R}_N, \\
    \mathfrak{R}_i = \big\{ \bm{y} \in \mathbb{R}^3 \vert ~ \bm{y} = R(\bm{s}_t)\bm{x}+\bm{\rho}(\bm{s}_t), ~A_i \bm{x} \preceq \bm{b}_i \big\},
  \end{gather*}
\end{subequations}
where $R(\bm{s}_t) \in \mathbf{SO}(3)$ and $\bm{\rho}(\bm{s}_t) \in \mathbb{R}^3$ are the rotation matrix and translation vector, respectively.
And $ A_i \in \mathbb{R}^{n_{r,i}\times 3}$ and $\bm{b}_i \in \mathbb{R}^{n_{r,i}}$ are parameters need to be determined according to the shape and related location of the $i$-th part of robots.
An example of quadrotor is shown in Fig. \ref{fig:geo_robot_obstacles}.

The geometry of obstacles is modeled by the union of a series of compact convex set, which is modeled by a polyhedron.
That is:
\begin{subequations}
  \begin{gather}
    \mathfrak{O} = \mathfrak{O}_1 \cup \mathfrak{O}_2 \cup \cdots \cup \mathfrak{O}_M, \\
    \mathfrak{O}_j = \big\{ \bm{y} \vert ~ C_j \bm{y} \preceq \bm{d}_j \big\},
  \end{gather}
\end{subequations}
where $ C_j \in \mathbb{R}^{n_{o,j}\times 3}$ and $\bm{d}_j \in \mathbb{R}^{n_{o,j}}$ are parameters need to be determined according to the shape and location of the $j$-th obstacles.
According to (\ref{eqn:poly_colli_free_conts}), the collision avoidance constraints between the $i$-th part of robot and the $j$-th obstacle at $t$-th step are obtained as:
\begin{subequations}
  \label{eqn:robot_colli_free_conts}
  \begin{gather}
    \bm{\lambda}_{ijt} \succeq 0, \qquad \bm{\mu}_{ijt} \succeq 0, \qquad \gamma_{ijt} \geq 0, \\
    -\bm{b}_i^T\bm{\lambda}_{ijt} + 1 = 0,                                   \\
    \mathcal{T}_{ijt}(\bm{s}_t,\bm{\mu}_{ijt},\gamma_{ijt}) = 0,      \\
    \mathcal{R}_{ijt}(\bm{s}_t,\bm{\lambda}_{ijt},\bm{\mu}_{ijt}) = 0,
  \end{gather}
\end{subequations}
where $\gamma_{ijt}$ is the slack variable, and $\mathcal{T}_{ijt}$ is the nonlinear function relates to translation vector,
\begin{equation}
  \begin{aligned}
    \mathcal{T}_{ijt}(\bm{s}_t,~ \bm{\mu}_{ijt},~ & \gamma_{ijt})                                                                        \\
    =                                             & ~ 1 + \big\{\bm{d}_j - C_j\bm{\rho}(\bm{s}_t)\big\}^T \bm{\mu}_{ijt} + \gamma_{ijt}.
  \end{aligned}
\end{equation}
And $\mathcal{R}_{ijt}$ is the nonlinear function relates to rotation matrix,
\begin{equation}
  \begin{aligned}
    \mathcal{R}_{ijt}(\bm{s}_t,\bm{\lambda}_{ijt},\bm{\mu}_{ijt})
    = A_i^T \bm{\lambda}_{ijt} + \big\{ C_j R(\bm{s}_t) \big\}^T \bm{\mu}_{ijt}.
  \end{aligned}
\end{equation}

The MPC generates the control input $\bm{u}_t$ by forecasting $T$-step future states and optimizing a cost function $\mathcal{F}$ under system, boundary and collision avoidance constraints.
The explicit form of cost function $\mathcal{F}$ depends on the requirements of task.
For instance, in trajectory tracking tasks, the tracking error and control efforts are take into account:
\begin{equation}
  \label{eqn:navigation_cost}
  \mathcal{F}(\{\bm{s}_t,\bm{u}_t\}) = \sum_{t=0}^{T} \bigg (\Vert \bm{s}_t - \tilde{\bm{s}}_t \Vert_{Q_s}^2 + \Vert \bm{u}_t \Vert_{Q_u}^2 \bigg),
\end{equation}
where $\tilde{\bm{s}}_t$ is the reference state, $Q_s$ and $Q_u$ are positive-definite matrixes with proper dimension.
The optimization problem of MPC is finally formulated as follows:
\begin{subequations}
  \label{eqn:MPC_opt_origin}
  \begin{align}
    \!\!\!\!\!\!\!\min_{\substack{\{\bm{s}_t,\bm{u}_t\},                                                                                                               \\ \{\bm{\lambda}_{ijt},\bm{\mu}_{ijt},\gamma_{ijt}\}.}}  &~~~~\sum_{t=0}^{T} \bigg (\Vert \bm{s}_t - \tilde{\bm{s}}_t \Vert_{Q_s}^2 + \Vert \bm{u}_t \Vert_{Q_u}^2 \bigg) \label{eqn:MPC_opt_origin:cost}        \\
    s.t. ~~~~ & ~~~~~~ \bm{s}_{t+1} = \bm{s}_t + f(\bm{s}_t,\bm{u}_t),~ \forall t,                                \label{eqn:MPC_opt_origin:sys_eqn}                   \\
              & ~~~~~~~~ \bm{s}_{min} \preceq \bm{s}_t \preceq \bm{s}_{max},~ \forall t,                                   \label{eqn:MPC_opt_origin:state}            \\
              & ~~~~~~~~ \bm{u}_{min} \preceq \bm{u}_t \preceq \bm{u}_{max},~ \forall t,                                   \label{eqn:MPC_opt_origin:control}          \\
              & \bm{\lambda}_{ijt} \succeq 0,~~ \bm{\mu}_{ijt} \succeq 0, ~~  \gamma_{ijt} \geq 0, ~\forall i,j,t,              \label{eqn:MPC_opt_origin:covar_1}     \\
              & ~~~~~-\bm{b}_i^T\bm{\lambda}_{ijt} + 1 = 0, ~\forall i,j,t,                                                \label{eqn:MPC_opt_origin:covar_2}          \\
              & ~~~~ \mathcal{T}_{ijt}(\bm{s}_t,\bm{\mu}_{ijt},\gamma_{ijt}) = 0, ~\forall i,j,t,                               \label{eqn:MPC_opt_origin:covar_trans} \\
              & ~~~~ \mathcal{R}_{ijt}(\bm{s}_t,\bm{\lambda}_{ijt},\bm{\mu}_{ijt}) = 0, ~\forall i,j,t,                    \label{eqn:MPC_opt_origin:covar_rot}
  \end{align}
\end{subequations}
where (\ref{eqn:MPC_opt_origin:sys_eqn}) is the system dynamic constraints, (\ref{eqn:MPC_opt_origin:state})-(\ref{eqn:MPC_opt_origin:control}) are the constraints on states and control input, and (\ref{eqn:MPC_opt_origin:covar_1})-(\ref{eqn:MPC_opt_origin:covar_rot}) are the collision avoidance constraints.

There are two challenges to solving the optimization problem (\ref{eqn:MPC_opt_origin}).
Firstly, the dimension of the optimization problem is related to the number of obstacles $M$.
When $M$ increases, the dimension of the optimization problem increases considerably, and hence real-time performance can not be guaranteed.
Secondly, the constraints (\ref{eqn:MPC_opt_origin:sys_eqn}), (\ref{eqn:MPC_opt_origin:covar_trans}) and (\ref{eqn:MPC_opt_origin:covar_rot}) are non-convex constraints in general, which makes the optimization problem hard to solve.

In this paper, we circumvent the first challenge by decomposing the original high-dimensional problem into several low-dimensional problems that can be solved in parallel.
Noting that for different $(i,j,t)$, the variables $(\bm{\lambda}_{ijt},~ \bm{\mu}_{ijt},~ \gamma_{ijt})$ are coupled indirectly with each other.
This property implies the potential to decompose optimization problem (\ref{eqn:MPC_opt_origin}) into several sub-problems only with $(\bm{\lambda}_{ijt},~ \bm{\mu}_{ijt},~ \gamma_{ijt})$ as variables.
To this end, the ADMM \cite{boyd2011distributed} is used to decompose the original problem (\ref{eqn:MPC_opt_origin}), which is detailed in in Subsection \ref{subsec:ADMM}.
We handle the second challenge by linearizing $f(\bm{s}_t,\bm{u}_t)$, $R(\bm{s}_t)$ and $\bm{\rho}(\bm{s}_t)$ with respect to $(\bm{s}_t,~\bm{u}_t)$.

\subsection{Problem Separation via ADMM}
\label{subsec:ADMM}
The ADMM handles the coupled constraints with augmented Lagrangian, and the augmented Lagrangian of (\ref{eqn:MPC_opt_origin}) is defined as:
\begin{equation}
  \label{eqn:ADMM_Lagrangian}
  \begin{aligned}
     & \mathcal{L}_{\sigma} (\{\bm{s}_t,~\bm{u}_t\},~\{\bm{\lambda}_{ijt},~\bm{\mu}_{ijt},\gamma_{ijt}\},~\{\zeta_{ijt},\bm{\xi}_{ijt}\})                                          \\
     & =  ~~\mathcal{F}(\{\bm{s}_t,\bm{u}_t\}) ~~ + ~~ IC_{0}(\{\bm{s}_t,\bm{u}_t\}) ~+                                                                                            \\
     & \frac{\sigma}{2} \sum_{i=1}^{N} \sum_{j=1}^{M} \sum_{t=0}^{T} \big \Vert \mathcal{T}_{ijt}(\bm{s}_t,~\bm{\mu}_{ijt},~\gamma_{ijt}) + \zeta_{ijt} \big \Vert_2^2  ~+         \\
     & \frac{\sigma}{2} \sum_{i=1}^{N} \sum_{j=1}^{M} \sum_{t=0}^{T} \big \Vert \mathcal{R}_{ijt}(\bm{s}_t,~\bm{\lambda}_{ijt},~\bm{\mu}_{ijt}) + \bm{\xi}_{ijt} \big \Vert_2^2 ~+ \\
     & \sum_{i=1}^{N} \sum_{j=1}^{M} \sum_{t=0}^{T} IC_{ijt}(\bm{\lambda}_{ijt},~\bm{\mu}_{ijt},~\gamma_{ijt} ),
  \end{aligned}
\end{equation}
where $\sigma > 0 $ is the penalty parameter.
$\zeta_{ijt}$ and $\bm{\xi}_{ijt}$ are the dual variables of ADMM.
$IC_{0}(\{\bm{s}_t,\bm{u}_t\})$ is the indicator function of constraints (\ref{eqn:MPC_opt_origin:sys_eqn}) - (\ref{eqn:MPC_opt_origin:control}).
That is, $IC_{0}(\{\bm{s}_t,\bm{u}_t\})=0$ if constraints (\ref{eqn:MPC_opt_origin:sys_eqn}) - (\ref{eqn:MPC_opt_origin:control}) are satisfied, and $IC_{0}(\{\bm{s}_t,\bm{u}_t\})=\infty$ otherwise.
$ IC_{ijt}(\bm{\lambda}_{ijt},~\bm{\mu}_{ijt},~\gamma_{ijt} ) $ is the indicator function of constraints (\ref{eqn:MPC_opt_origin:covar_1}) - (\ref{eqn:MPC_opt_origin:covar_2}).

The ADMM is an iterative algorithm to solve optimization problems.
In each iteration, the ADMM updates the optimization variables via solving
sub-problems.
Specially, for augmented Lagrangian (\ref{eqn:ADMM_Lagrangian}), there are three steps.
In $(k+1)$-th iteration, the first step is the updates of dual variables:
\begin{equation}
  \label{eqn:ADMM_dual_updates}
  \begin{aligned}
    \!\!\!\!\!\!\!\!\! & ~~~\{\bm{\lambda}_{ijt}^{k+1}, \bm{\mu}_{ijt}^{k+1}, \gamma_{ijt}^{k+1}\} =                                                                                                                                                \\
    \!\!\!\!\!\!\!\!\! & \mathop{\arg\min}_{\{\bm{\lambda}_{ijt},\bm{\mu}_{ijt},\gamma_{ijt}\}}  \!\!\!\!\! \mathcal{L}_{\sigma} (\{\bm{s}_t^k,\bm{u}_t^k\},\{\bm{\lambda}_{ijt},\bm{\mu}_{ijt},\gamma_{ijt}\},\{\zeta_{ijt}^k,\bm{\xi}_{ijt}^k\}).
  \end{aligned}
\end{equation}
The second step is the updates of states and control inputs:
\begin{equation}
  \label{eqn:ADMM_state_updates}
  \begin{aligned}
     & \{\bm{s}_t^{k+1},\bm{u}_t^{k+1}\} =                                                                                                                                                    \\
     & \!\!\mathop{\arg\min}_{\{\bm{s}_t,\bm{u}_t\}}\mathcal{L}_{\sigma} (\{\bm{s}_t,\bm{u}_t\},\{\bm{\lambda}_{ijt}^k,\bm{\mu}_{ijt}^k,\gamma_{ijt}^k\},\{\zeta_{ijt}^k,\bm{\xi}_{ijt}^k\}).
  \end{aligned}
\end{equation}
The third step is the updates of dual variables of ADMM:
\begin{equation}
  \label{eqn:ADMM_residue_updates}
  \begin{aligned}
     & ~~~~\zeta_{ijt}^{k+1} = \zeta_{ijt}^k +  \mathcal{T}_{ijt}(\bm{s}_t^k,~\bm{\mu}_{ijt}^k,~\gamma_{ijt}^k),~ \forall i,j,t,             \\
     & ~~~~\bm{\xi}_{ijt}^{k+1} = \bm{\xi}_{ijt}^k + \mathcal{R}_{ijt}(\bm{s}_t^k,~\bm{\lambda}_{ijt}^k,~\bm{\mu}_{ijt}^k),~ \forall i,j,t .
  \end{aligned}
\end{equation}

\begin{figure}[!t]
  \centering
  \includegraphics[width=0.95\hsize]{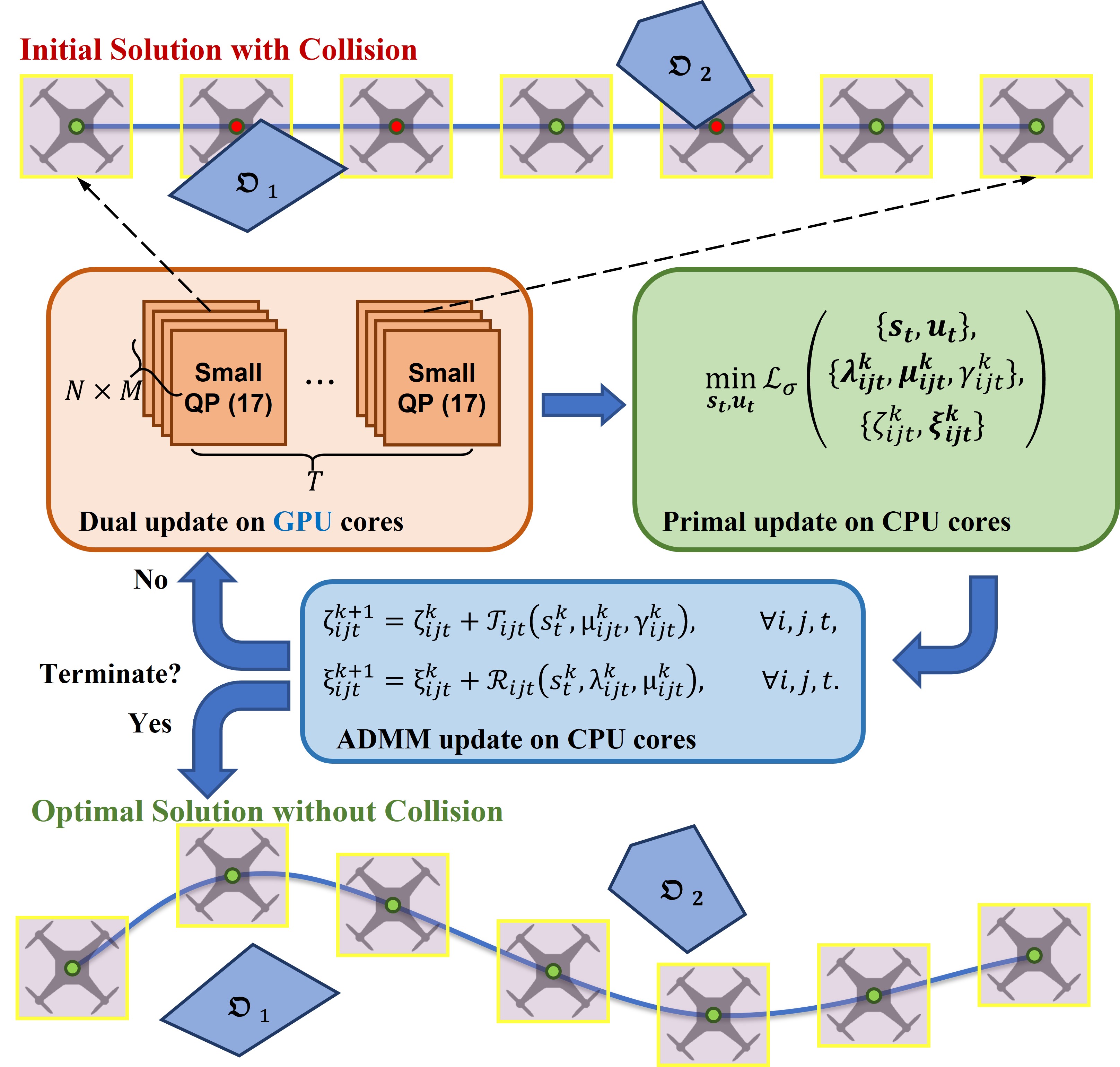}
  \caption{The overview of our optimization process.}
  \label{fig:overview}
\end{figure}

The stopping criteria terminating the iterative procedure is given by the
following conditions,
\begin{gather}
  \!\!\sum_{i}\! \sum_{j}\! \sum_{t}\! \bigg(\Vert \zeta_{ijt}^{k+1} \!-\!\zeta_{ijt}^{k} \Vert_2^2 \!+ \!\Vert \bm{\xi}_{ijt}^{k+1} -\bm{\xi}_{ijt}^{k} \Vert_2^2 \bigg) \! \leq \! \epsilon_{pri}, \\
  \!\!\sum_{i}\! \sum_{j}\! \sum_{t}\! \bigg(\Vert \bm{\lambda}_{ijt}^{k+1} \!\!- \! \bm{\lambda}_{ijt}^{k} \Vert_2^2 \!+\! \Vert \bm{\mu}_{ijt}^{k+1}\!\! -\! \bm{\mu}_{ijt}^{k} \Vert_2^2 \bigg) \!\leq\! \epsilon_{dual},
\end{gather}
where the $\epsilon_{pri} > 0$ and $\epsilon_{dual} > 0$ are the constants that control the feasibility of constants (\ref{eqn:MPC_opt_origin:covar_trans})-(\ref{eqn:MPC_opt_origin:covar_rot}) and optimality of solution, respectively.
To trade off between feasibility and optimality, $\epsilon_{pri}$ and $\epsilon_{dual}$ need to be chosen empirically.
The overview of our optimization process is shown in Fig. \ref{fig:overview}.

Noting that the variables with superscript are regarded as constant value in
(\ref{eqn:ADMM_dual_updates})-(\ref{eqn:ADMM_residue_updates}). Consequently,
the optimization problem (\ref{eqn:ADMM_dual_updates}) is a QP with linear constraints.
Furthermore, optimization problem (\ref{eqn:ADMM_dual_updates}) is inherently decoupled for variables with different $(i,j,t)$.
Consequently, optimization problem (\ref{eqn:ADMM_dual_updates}) can be separated into a series of low-dimensional QPs that can be solved in parallel with GPUs.
We will discuss about the parallel solving method in \ref{subsec:GPU}.

The second challenge mentioned before appears in the optimization problem
(\ref{eqn:ADMM_state_updates}).
That is, the optimization problem (\ref{eqn:ADMM_state_updates}) is a non-linear programming (NLP) due to the nonlinearity of $f(\bm{s}_t,\bm{u}_t)$, $R(\bm{s}_t)$ and $\bm{\rho}(\bm{s}_t)$.
In this paper, the NLP is optimized with sequential quadratic programming (SQP).
More precisely, the NLP is approximated by a series of QPs.
To enhance real-time performance, we only solve the QP once in each iteration of ADMM.

\subsection{Solving Multiple QPs with GPU in Parallel}
\label{subsec:GPU}

The sub-problems in (\ref{eqn:ADMM_dual_updates}) are equivalent to the following QP,
\begin{subequations}
  \label{eqn:const_least_square}
  \begin{align}
    \min_{\bm{y}_{ijt}} ~~ & ~~ \frac{1}{2} \Vert K_{ijt}^T ~\bm{y}_{ijt} + \bm{b}_{ijt} \Vert_2^2 \label{eqn:const_least_square:cost_function}         \\
    s.t. ~~                & ~~~~~ \bm{\kappa}_{ijt}^T ~\bm{y}_{ijt} = \bm{\eta}_{ijt},              \label{eqn:const_least_square:equality_constraint} \\
    ~~~~                   & ~~~~~~~~~~~ \bm{y}_{ijt} \succeq \bm{0} \label{eqn:const_least_square:non-negative_constraint},
  \end{align}
\end{subequations}
where the parameters are detailed as,
\begin{subequations}
  \begin{gather}
    \bm{y}_{ijt} =
    \begin{bmatrix}
      \bm{\lambda}_{ijt}^T & \bm{\mu}_{ijt}^T & \gamma_{ijt}
    \end{bmatrix}^T \in \mathbb{R}^{n_r+n_o+1},\\
    K_{ijt} =
    \begin{pmatrix}
      \bm{0}                    & A_i     \\
      \bm{d}_j - C_j\bm{\rho}_t & C_j R_t \\
      1                         & \bm{0}
    \end{pmatrix} \in \mathbb{R}^{(n_r+n_o+1)\times 4},\\
    \bm{b}_{ijt} =
    \begin{pmatrix}
      1 + \zeta_{ijt} \\
      \bm{\xi}_{ijt}
    \end{pmatrix}\in \mathbb{R}^{4}, \\
    \bm{\kappa}_{ijt} = \begin{pmatrix}
      \bm{b}_{i} \\ \bm{0}
    \end{pmatrix} \in \mathbb{R}^{n_r+n_o+1}, \quad
    \bm{\eta}_{ijt} = 1.
  \end{gather}
\end{subequations}

There are $(N \times M \times T)$ QPs in total, which is the bottleneck of the ADMM updates.
We avoid this bottleneck by solving these QPs in parallel.
More precisely, the parallel version of Lemke's algorithm \cite{kipfer2007lcp} are implemented on GPU and used to solve these problem.
To use Lemke's algorithm, the QPs (\ref{eqn:const_least_square}) should be converted into linear complementarity problems (LCPs).
The conversion are separated in two steps.

For simplicity, the subscripts $(i,j,t)$ are  omitted in the remainder of
this subsection.
The first step is to convert the linear equality constraint (\ref{eqn:const_least_square:equality_constraint}) into linear inequality
constraint.
The variables and parameters are partitioned as:
\begin{equation}
  \bm{y} = \begin{pmatrix}
    \bm{y}_e \\\bm{y}_u
  \end{pmatrix},~
  K = \begin{pmatrix}
    K_e \\ K_u
  \end{pmatrix},~
  \bm{\kappa} = \begin{pmatrix}
    \bm{\kappa}_e \\\bm{\kappa}_u
  \end{pmatrix},
\end{equation}
where $\bm{\kappa}_e$ is an invertible square matrix with proper dimension.
According to (\ref{eqn:const_least_square:equality_constraint}), we have:
\begin{equation}
  \bm{y}_e = \bm{\kappa}_e^{-T}(\bm{\eta}-\bm{\kappa}_u^{T}\bm{y}_u).
\end{equation}
Submitting the above equation into (\ref{eqn:const_least_square}), we have the following new QP,
\begin{subequations}
  \label{eqn:const_least_square_ineq}
  \begin{align}
    \min_{\bm{y}_u} ~~ & ~\frac{1}{2} \Vert \tilde{K}^T \bm{y}_u + \tilde{\bm{b}} \Vert_2^2 \label{eqn:const_least_square_ineq:cost_function}                   \\
    s.t. ~~            & ~~~~ \tilde{\bm{\kappa}}^T ~\bm{y}_u \preceq \tilde{\bm{\eta}},              \label{eqn:const_least_square_ineq:inequality_constraint} \\
    ~~~~               & ~~~~~~~~~ \bm{y}_u \succeq \bm{0} \label{eqn:const_least_square_ineq:non-negative_constraint},
  \end{align}
\end{subequations}
where the new parameters are detailed as,
\begin{subequations}
  \begin{gather}
    \tilde{K} = K_u - \bm{\kappa}_u \bm{\kappa}_e^{-1} K_e, \quad
    \tilde{\bm{b}} =  \bm{b} + (\bm{\kappa}_e^{-1} K_e)^T \bm{\eta},\\
    \tilde{\bm{\kappa}} = \bm{\kappa}_u \bm{\kappa}_e^{-1}, \quad \tilde{\bm{\eta}} = \bm{\kappa}_e^{-T}\bm{\eta}.
  \end{gather}
\end{subequations}

The second step is to convert the new QP (\ref{eqn:const_least_square_ineq})
into LCP. The Karush-Kuhn-Tucker (KKT) conditions of
(\ref{eqn:const_least_square_ineq}) is,
\begin{subequations}
  \label{eqn:kkt_conditions}
  \begin{gather}
    \tilde{K}\tilde{K}^T \bm{y}_u + \tilde{K} \tilde{\bm{b}} + \tilde{\bm{\kappa}} \bm{\phi} -\bm{\psi} = \bm{0},\\
    \tilde{\bm{\kappa}}^T \bm{y}_u -\bm{\tilde{\eta}} + \bm{\varepsilon} = \bm{0},\\
    \bm{\phi}^T  \bm{\varepsilon}= 0,\quad \bm{\psi}^T \bm{y}_u = 0,\\
    \bm{\phi}  \succeq \bm{0},\quad \bm{\varepsilon}  \succeq \bm{0}, \quad \bm{y}_u \succeq \bm{0}, \quad \bm{\psi} \succeq \bm{0},
  \end{gather}
\end{subequations}
where $\bm{\phi}$ and $\bm{\psi}$ are the dual variables of inequality constraints (\ref{eqn:const_least_square_ineq:inequality_constraint}) and (\ref{eqn:const_least_square_ineq:non-negative_constraint}), respectively.
And $\bm{\varepsilon}$ is the slack variable of (\ref{eqn:const_least_square_ineq:inequality_constraint}).
The KKT conditions (\ref{eqn:kkt_conditions}) is equivalent to the following LCP:
\begin{equation}
  \label{eqn:linear_complementarity_problem}
  \begin{aligned}
    \bm{w} = M\bm{z} + \bm{q}, & \quad \bm{w}^T \bm{z} =0,    \\
    \bm{w} \succeq \bm{0},     & \quad \bm{z} \succeq \bm{0},
  \end{aligned}
\end{equation}
where the variables and parameters are detailed as,
\begin{equation}
  \begin{aligned}
    \bm{z} = \begin{pmatrix}
               \bm{y_u} \\ \bm{\phi}
             \end{pmatrix},                  & \quad
    \bm{w} = \begin{pmatrix}
               \bm{\psi} \\ \bm{\varepsilon}
             \end{pmatrix},            \\
    M = \begin{pmatrix}
          \tilde{K} \tilde{K}^T & \tilde{\kappa} \\
          -\tilde{\kappa}^T     & \bm{0}
        \end{pmatrix}, & \quad
    \bm{q} = \begin{pmatrix}
               \tilde{K} \tilde{\bm{b}} \\ \tilde{\eta}
             \end{pmatrix}.
  \end{aligned}
\end{equation}
The conversion is completed.

\subsection{Computational Complexity}
There are two levels of acceleration to solve (\ref{eqn:ADMM_dual_updates}) with GPU.
The first level is the acceleration of solving a single sub-problem.
Denotes $n$ as the dimension of $\bm{z}$, the time complexity of solving QP with single thread CPU program is $O(n^3)$.
On the contrary, the Lemke's algorithm typically requires $O(n)$ iterations to terminate \cite{lloyd2005fast}, and the time complexity of each iteration is $O(n)$ when implemented with GPU \cite{kipfer2007lcp}.
As a result, the time complexity for solving one sub-problem with GPU is of order $O(n^2)$.

The second level is the acceleration of solving multiple sub-problems.
There are $(N \times M \times T)$ sub-problems in total.
These sub-problems can be solved simultaneously using multi-threads technique of both CPU and GPU.
A CPU generally has 4-16 threads, on the contrary, a GPU can generally handle
thousands of threads within one clock cycle.
Hence, GPUs outperform CPUs when the number of sub-problems is large.
For CPU, the time complexity for solving all sub-problems of (\ref{eqn:ADMM_dual_updates}) is of order,
\begin{equation}
  O\bigg( ceil(\frac{NMT}{K_{CPU}}) \cdot (n_{max} + 1)^3\bigg),
\end{equation}
where $ceil(x)$ is the function that computes the smallest integer that is greater than or equal to $x$.
$K_{CPU}$ is the number of threads of CPU, and $n_{max} = \max\limits_{i,j}~ (n_{r,i} + n_{o,j})$ is the maximum dimension of sub-problems.
For GPU, the time complexity for solving all sub-problems of (\ref{eqn:ADMM_dual_updates}) is of order:
\begin{equation}
  O \bigg( ceil(\frac{NMT}{ceil(K_{GPU}/(n_{max}+1))}) \cdot (n_{max} + 1)^2 \bigg),
\end{equation}
where $K_{GPU}$ is the number of threads of GPU.
It is worth mentioning that $K_{GPU} \gg K_{CPU}$ in general.

\section{Simulation and Benchmark Comparisons}
\label{sec:simulation}

In this section, high-fidelity simulations on quadrotor are conducted to illustrate the effectiveness of our algorithm.
We focus on the collision-free trajectory tracking problem of quadrotors.
More precisely, the quadrotor should move as close as possible to a reference trajectory while avoiding collisions with obstacles.

The proposed method is also compared with the following state-of-the-art optimization based method.
\begin{enumerate}
  \item Optimization-based collision avoidance (OBCA) \cite{zhang2020optimization}, which use sign distance-based collision avoidance constraints and solve the MPC problem via non-convex optimization method like quasi-newton method or sequential convex programming.
        The OBCA method is generally regarded as an offline planner.
  \item Regularized dual alternating direction method of
        multiplier (RDA) \cite{han2023rda}, which use sign distance-based collision avoidance constraints.
        RDA formulate the update of dual variables as SOCPs.
        And these SOCPs are solved using multi-threads technique of CPU.
\end{enumerate}

All these method are evaluated on an embedded platform called Nvidia Jetson Xavier NX.

\subsection{High-Fidelity Simulation}
\begin{figure}[t!]
  \centering
  \includegraphics[width=1.0\hsize]{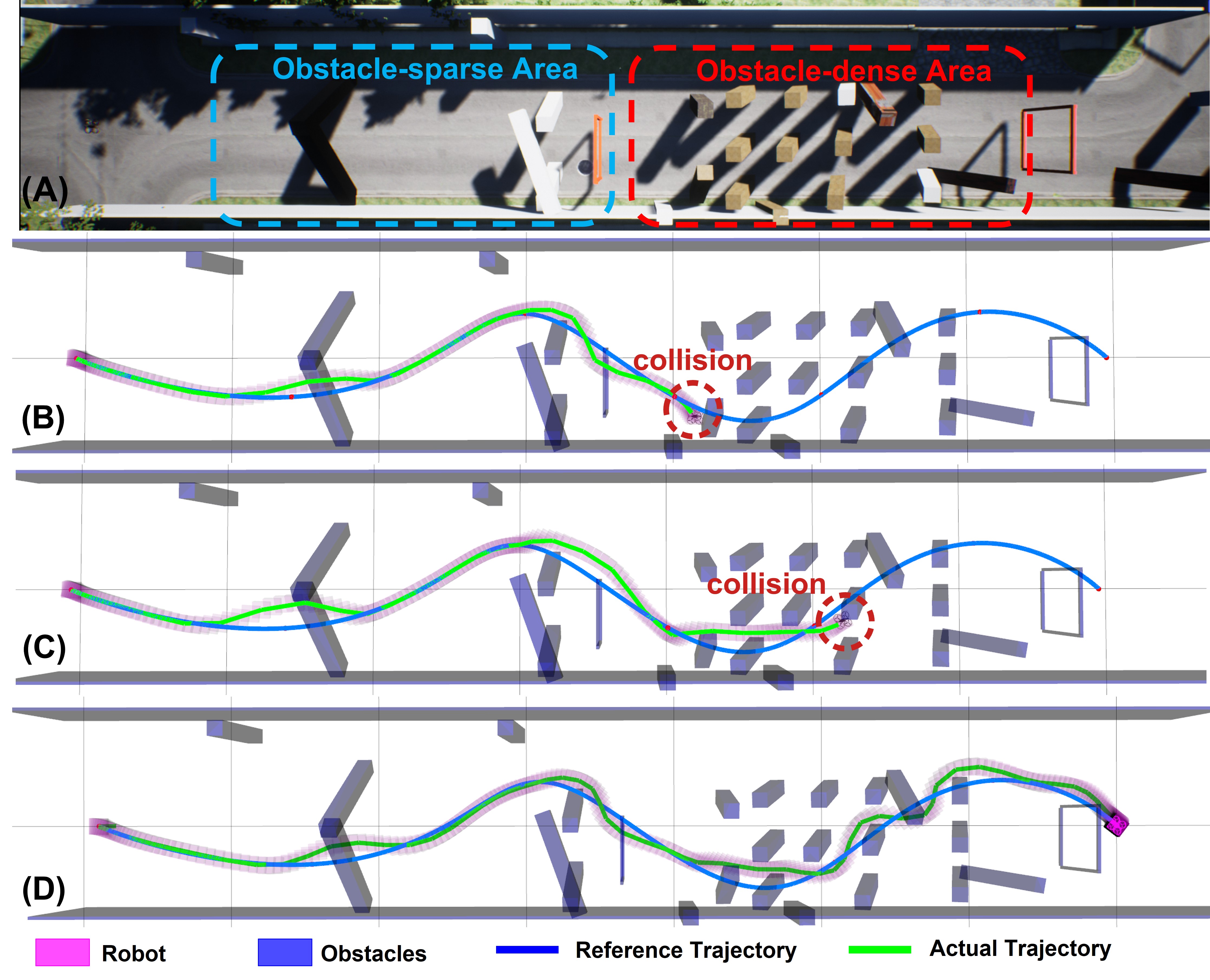}
  \caption{One case in the simulation. (A) The environment in AirSim. (B) The geometry of obstacles, the geometry of the quadrotor, the reference trajectory and the actual trajectory during the navigation with OBCA. (C) The navigation with RDA. (D) The navigation with the proposed method.}
  \label{fig:airsim_trace}
  \vspace{-0.5em}
\end{figure}

A high-fidelity quadrotor simulator called AirSim \cite{shah2018airsim} is used to verify the effectiveness of the proposed method under practical sensor and dynamic constraints.
The simulator is running on a laptop equipped with an Intel Core i9-12700 CPU and a GeForce RTX 3070 Ti GPU.
The simulation scenario is of size $70m \times 20m \times 6m$ and contains 32 obstacles in total.
In particular, the scenario contains an obstacle-sparse area and an obstacle-dense area as shown in Fig. \ref{fig:airsim_trace}(A).

For all method, the discrete time $\Delta t = 0.1s$, the prediction horizon $T=16$.
The penalty parameter $\sigma$ of RDA and proposed method is set to $300$.
The initial position of quadrotor is $(0,0,1)$, the goal position is $(0,70,1)$, the duration of reference trajectories is set to $25s$.
The reference trajectories are generated by interpolating initial position, goal position and $5$ random waypoints.
During the navigation, the quadrotor can only sense the obstacles within $20m \times 20m \times 6m$.

One particular case of different methods is shown in Fig. \ref{fig:airsim_trace}(B)-(C).
The quadrotor successfully passes through the obstacle-sparse area with all three methods.
However, the OBCA and RDA methods lead to a collision in the obstacle-dense area.
Table \ref{tbl:benchmark_tracking} reports the success rate, navigation times, and navigation cost $\mathcal{F}$ (\ref{eqn:navigation_cost}) of 100 trials.
The three methods achieve similar performance in navigation time and navigation cost, while the proposed method outperforms OBCA and RDA in success rate.

\begin{table}[htbp]
  \tabcolsep=0.175cm
  \centering
  \caption{Benchmark Comparison of Simulations}
  \label{tbl:benchmark_tracking}
  \begin{tabular}{*{8}{c}}
    \toprule
    \multirow{2}{*}{Method}           & \multirow{2}{*}{\makecell[c]{success                                                                                                       \\ rate}} & \multicolumn{3}{c}{Navigation Time$^\star$} & \multicolumn{3}{c}{Navigation Cost$^\star$ $\mathcal{F}$}                           \\    \cmidrule(lr){3-5} \cmidrule(lr){6-8}
                                      &                                      & min            & max            & mean           & min            & max            & mean           \\ \midrule
    OBCA \cite{zhang2020optimization} & 35\%                                 & \textbf{26.6s} & 32.7s          & \textbf{28.3s} & 380.7          & \textbf{532.6} & 483.9          \\
    RDA  \cite{han2023rda}            & 58\%                                 & 27.2s          & \textbf{32.1s} & 28.5s          & 348.1          & 576.1          & 492.6          \\
    Proposed                          & \textbf{92\%}                        & 26.8s          & 33.8s          & 28.7s          & \textbf{322.4} & 558.2          & \textbf{458.7} \\
    \bottomrule
  \end{tabular}
  \begin{tablenotes}
    \item $^\star$ Only the data of success trials are taken into account.
  \end{tablenotes}
  \vspace{-0.5em}
\end{table}

\begin{figure}[htbp]
  \centering
  \subfigure[Average computation time of one control loop with different methods when prediction horizon $T=16$.]{
    \includegraphics[clip, trim={1.0cm 0.0cm 1.0cm 0.9cm}, width=1.0\linewidth]{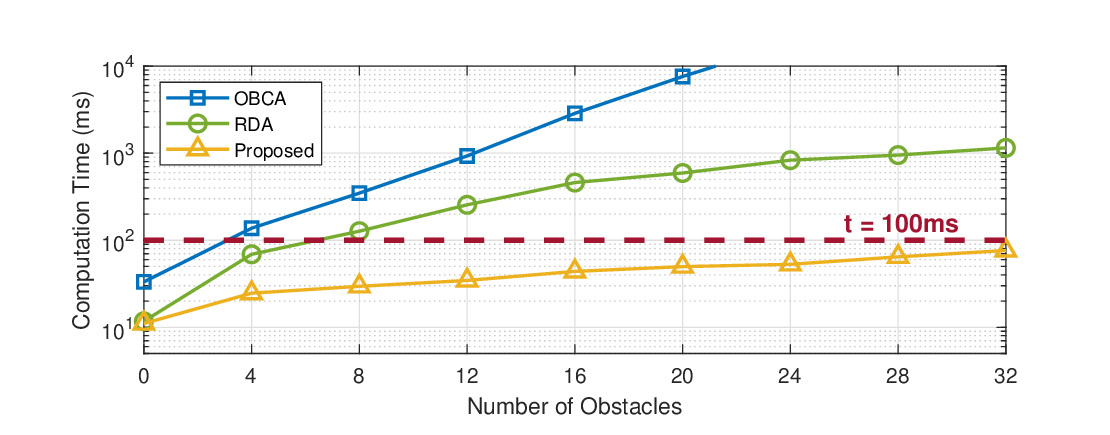}
    \label{fig:time_logmap_all}}
  \subfigure[Dual updates of RDA.]{
    \includegraphics[width=0.46\linewidth]{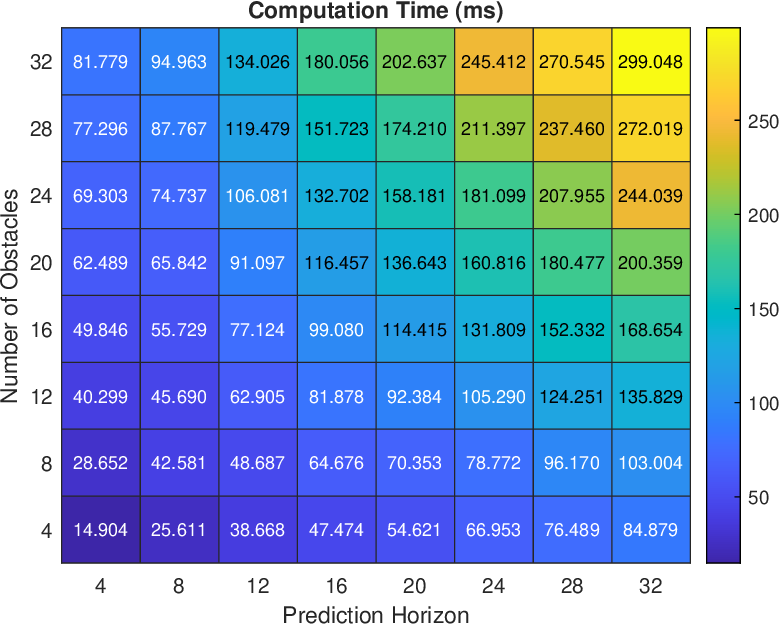}
    \label{fig:time_heatmap:rda}}
  \subfigure[Dual updates of our method.]{
    \includegraphics[width=0.46\linewidth]{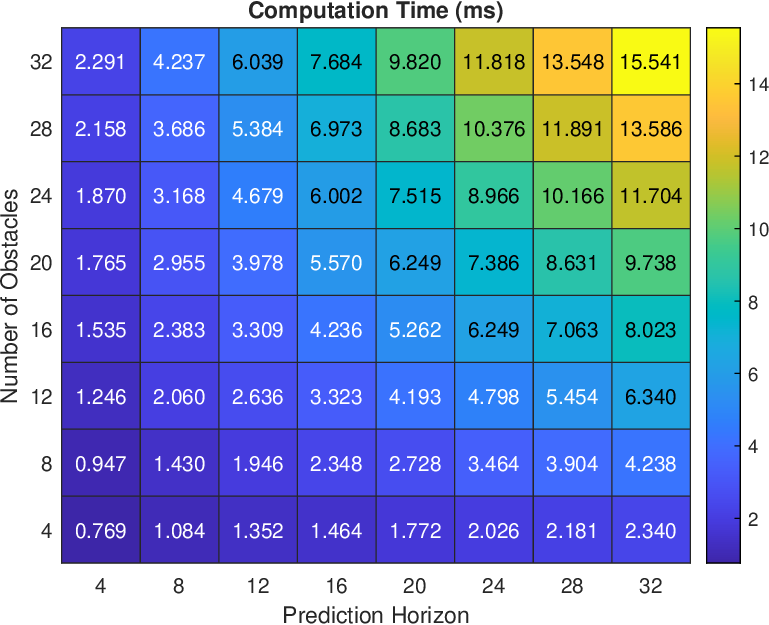}
    \label{fig:time_heatmap:proposed}}
  \caption{Comparisons of computation time. }
  \label{fig:computation_time}
  \vspace{-0.5em}
\end{figure}

To further explain the result in Table \ref{tbl:benchmark_tracking}, the computational time of different methods is summarized in Fig. \ref{fig:computation_time}.
The OBCA method solve the optimization problem (\ref{eqn:MPC_opt_origin}) directly, whose dimension is proportional to the number of obstacles $M$.
Consequently, the computational time of OBCA method increase dramatically with $M$.
As shown in Fig. \ref{fig:time_logmap_all}, the OBCA method can not generate safe trajectories within $0.1s$ in the obstacle-dense area, which causes collisions.
The RDA method and the proposed method both solve the optimization problem (\ref{eqn:MPC_opt_origin}) with ADMM.
The main difference between the two methods is that the RDA method uses distance-based collision detection, which leads to solving a series of SOCPs to update dual variables.
These SOCPs are solved in parallel via the thread pool technique of CPUs.
On the contrary, the proposed method uses scale-based collision detection, which leads to solving a series of QPs to update dual variables as shown in Section \ref{subsec:ADMM}.
Moreover, these QPs are solved in parallel via GPUs.
We compare the computation times of dual updates between the RDA method and the proposed method on an embedded platform called Xavier NX, which has 6 CPU cores and 384 GPU cores.
The results are shown in Fig. \ref{fig:time_heatmap:rda} and Fig. \ref{fig:time_heatmap:proposed}.
It can be seen that the proposed method outperforms the RDA method by orders of magnitudes.
As a result, the proposed method can still generate safe trajectories timely in obstacle-dense area.

\section{Conclusion}
This paper proposed a GPU-accelerated optimization framework for the collision avoidance problem.
With the help of scale-based collision detection and ADMM, the optimization problem is separated into multiple low-dimensional QPs, which can be solved in parallel with GPUs.
High-fidelity simulations on quadrotors are conducted to show the effectiveness of the proposed framework.
The simulation results have shown that the proposed framework significantly reduces the computational time of optimization problems.
Moreover, the benchmark comparisons indicate that the proposed framework performs better on embedded platforms than OBCA and RDA.

\balance
\footnotesize{
  \bibliographystyle{IEEEtranN}
  \bibliography{IEEEabrv, papers.bib}
}

\addtolength{\textheight}{-12cm}
\end{document}